\documentclass[letterpaper, 10.2 pt, conference]{ieeeconf}  

\usepackage[utf8]{inputenc}
\usepackage{graphicx}
\usepackage{epstopdf}
\usepackage{etoolbox}
\usepackage{cite}
\usepackage{picinpar}
\usepackage{amsmath}
\usepackage{url}
\usepackage{flushend}
\usepackage{textcomp}
\usepackage{bm}
\usepackage{makecell}
\usepackage{comment}

\IEEEoverridecommandlockouts                              

\title{\LARGE \bf
Control of Soft Pneumatic Actuators with Approximated\\ Dynamical Modeling
}

\author{Wu-Te Yang$^{1}$, Burak Kürkçü$^{2}$, Motohiro Hirao$^{3}$, Lingfeng Sun$^{1}$, Xinghao Zhu$^{1}$\\Zhizhou Zhang$^{1}$, Grace X. Gu$^{1}$, and Masayoshi Tomizuka$^{1}$
\thanks{$^{1}$The authors are with the Department of Mechanical Engineering, University of California, Berkeley, USA. {\tt\small wtyang, lingfengsun, zhuxh, zz\_zhang, ggu, tomizuka@berkeley.edu}}%
\thanks{$^{2}$The author is with the Department of Computer Engineering, Hacettepe University Turkey, and also is a Research Scholar at the University of California, Berkeley, USA {\tt\small bkurkcu@berkeley.edu}} 
\thanks{$^{3}$The author is a visiting fellowhsip with the NSK Ltd., Japan {\tt\small hirao@berkeley.edu}}%
}

\begin{document}

\maketitle
\thispagestyle{empty}
\pagestyle{empty}

\begin{abstract}

This paper introduces a full system modeling strategy for a syringe pump and soft pneumatic actuators(SPAs). The soft actuator is conceptualized as a beam structure, utilizing a second-order bending model. The equation of natural frequency is derived from Euler's bending theory, while the damping ratio is estimated by fitting step responses of soft pneumatic actuators. Evaluation of model uncertainty underscores the robustness of our modeling methodology. To validate our approach, we deploy it across four prototypes varying in dimensional parameters. Furthermore, a syringe pump is designed to drive the actuator, and a pressure model is proposed to construct a full system model. By employing this full system model, the Linear-Quadratic Regulator (LQR) controller is implemented to control the soft actuator, achieving high-speed responses and high accuracy in both step response and square wave function response tests. Both the modeling method and the LQR controller are thoroughly evaluated through experiments. Lastly, a gripper, consisting of two actuators with a feedback controller, demonstrates stable grasping of delicate objects with a significantly higher success rate.

\end{abstract}

\section{INTRODUCTION}

Soft robots, constructed from highly elastic materials, have demonstrated superior performance compared to rigid robots, particularly in scenarios involving unknown environments~\cite{c1, c2}, ensuring the safety of human-robot collaboration~\cite{c3}, and handling fragile objects in the food and agriculture industry~\cite{c4, c25}. These soft robots are typically driven by soft actuators, such as electroactive polymers, cable-driven systems, shape memory alloys, and soft pneumatic actuators~\cite{c5,zhang2022efficient}. Among these options, soft pneumatic actuators are widely used~\cite{c5, c6} due to their lightweight, cost-effectiveness, and high power density. However, their complex geometric shapes and pressure-driven nature present challenges for dynamical modeling and subsequent control~\cite{c7, c8}, particularly when implementing model-based control methods~\cite{c26}. 

\begin{figure}[t]
    \centering
    \includegraphics[width=190pt]{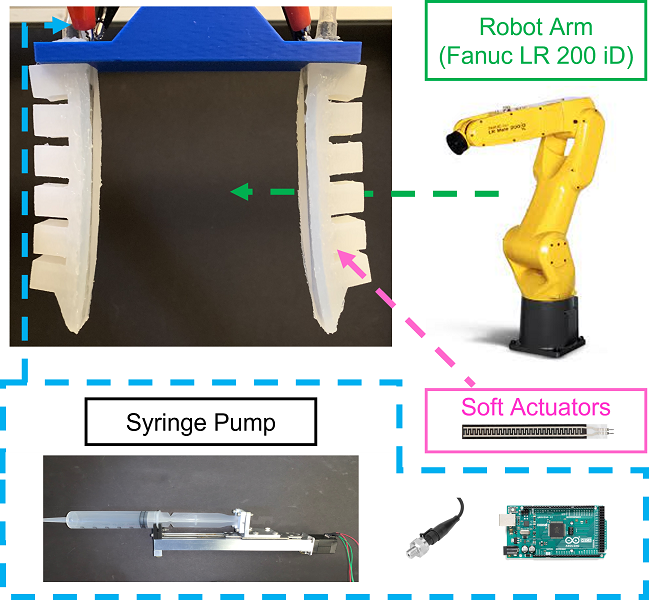}
    \caption{The framework of this research. A full system modeling method is proposed to model soft actuators and syringe pump. A soft gripper made of two actuators driven by the syringe pump. The designed optimal controller is programmed inside the microcontroller. The soft gripper is deployed in an industrial robot arm.}
    \vspace{-0.2in}
    \label{fig: 1}
\end{figure}
Recently, a couple of works~\cite{c8, c14 ,c15, c18, c19} modeled the soft pneumatic as the second-order dynamic system by fitting systems' responses to determine the damping ratio and natural frequency. Unlike the analytical method, curve fitting methods rely on experimental data to get accurate system parameters. In addition, the full system model which includes models of the pneumatic system and soft actuator, is important to the controller design. While some pressure models of air tanks with solenoid valves~\cite{c8, c15, c20} have been developed to construct a comprehensive system model, controlling the solenoid valve posed challenges. It is driven by pulse-width modulation signals, whereas the syringe pump is powered by a rotary motor, making it comparatively easier to manage.
Last but not least, previous works implemented the adaptive controllers~\cite{c14, c32} and nonlinear controllers~\cite{c18, c19, c8, c32} to deal with the nonlinearity of soft robots. However, the optimal controllers minimizing the errors of system states which is especially important for grasping works were barely mentioned to address soft actuators.

The aim of the paper is to build a full system model, including models of a syringe pump and soft actuator, for a soft pneumatic actuator. Then, an optimal controller is designed for the system to achieve fast and precise responses. Firstly, we approximate the complex-shaped structure of the pneumatic actuator as a cantilever beam, as depicted in Fig.~\ref{fig: 2} (a). The approximated structure is analyzed by Euler's bending beam theory~\cite{c22} to determine the natural frequency analytically. The damping ratio is obtained by fitting the step responses of the actuator. Secondly, we design a syringe pump by combining a linear actuator and a commercial syringe to drive the soft actuator. Additionally, the pressure model of the syringe pump is developed to predict the pressure changes inside the soft actuator. The pressure model and the actuator's model form a full model for controller design. Lastly, the Linear-Quadratic Regulator (LQR) controller is implemented to the full model to achieve high-speed responses and reduce control errors.


The remainder of this paper is organized as follows. Section II compares related works with the proposed method. Section III describes the full system modeling for the SPA and the syringe pump. Section IV discusses the controller design and its simulation. Section V demonstrates the experimental results of the soft pneumatic actuator. Section VI discusses the experimental results and concludes the work.

\section{Related Works}
\subsubsection{Soft Actuator Modeling}Various modeling methods are proposed for soft actuators. Bending models for soft actuators are developed by using the piece-wise constant curvature method~\cite{c9}, Euler's bending beam theory~\cite{c10}, Cosserat rod model~\cite{c11, c24}, and Lagrange equation~\cite{c12, c16}. Recent works modeled the soft actuators as second-order systems by curve fitting or least square methods~\cite{c8, c14 ,c15, c18, c19}. Instead, we propose an analytical method to compute the natural frequency of soft actuator. The approach serves as an alternative way to obtain the natural frequency. 
\subsubsection{Syringe Pump Modeling}
 Some syringe pumps design~\cite{syringe, s1} were proposed and applied to control soft robots; however, they mainly focused on the performance such as control accuracy. They did not develop the pressure model of the syringe pump which had an influence on the controller design. Thus, the pressure dynamic model of the syringe pump is developed to build a full model for later controller design. 
\subsubsection{Controller Design}The controllers can be developed for the soft pneumatic actuator based on the dynamic model. The straightforward strategies include open-loop control~\cite{c13} and closed-loop controller (PID controller)~\cite{c11, c17, c24, c27}. However, the fundamental controllers are primarily used for preliminary simulations. On the other hand, the model reference adaptive controller is adopted to handle the model uncertainties caused by the soft materials~\cite{c14, c32}. Other advanced controllers such as the sliding mode controller~\cite{c18, c19}, feedback linearization controller~\cite{c8, c20}, and backstepping controller~\cite{c15} are utilized to deal with the unpredictable soft actuators. The state feedback controllers~\cite{c21, c23} placing poles at desired locations could also achieve desired responses. Nonetheless, the LQR controller which minimizes the state errors and control input is rarely applied to handle the soft robots. Thus, we implement the LQR controller in the system to attain rapid responses, higher accuracy, and synchronized control across multiple fingers. 

Overall, we seek to produce an alternative and useful full system modeling and controller design approach for the SPAs driven by the syringe pump.

\begin{figure}[http]
    \centering
    \vspace{0.1in}
    \includegraphics[width=190pt]{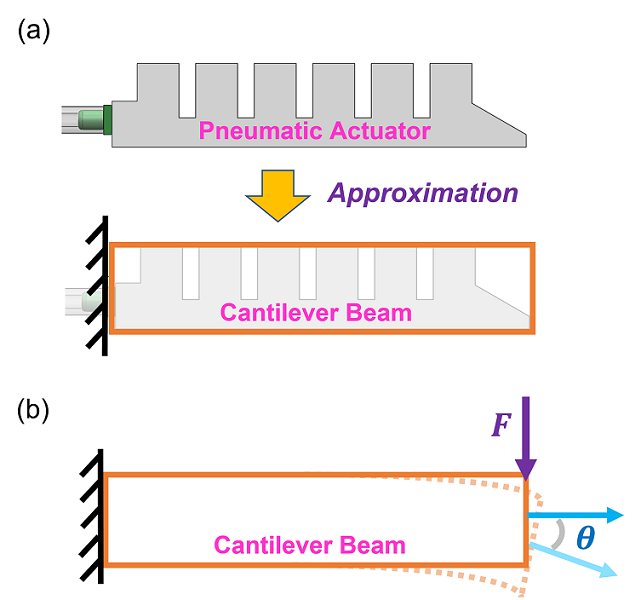}
    \caption{(a) The nonlinear structure of soft pneumatic actuator is approximated by a cantilever beam. (b) The approximated structure of the soft actuator generates bending angle $\theta$ with a load $F$. }
    \vspace{-0.15in}

    \label{fig: 2}
\end{figure}

\section{System Modeling}
In this section, we build the dynamical model for both the soft actuator and the pneumatic control system. The data-driven models are also constructed to verify the effectiveness of the proposed modeling method.


\subsection{Modeling of the Soft Actuator}

SPAs consist of numerous discrete chambers, and these irregular structures make them hard to model. Currently, researchers might count on curve fitting method to get the dynamic model\cite{c8, c14 ,c15, c18, c19}. This paper approximates the soft pneumatic actuator as a cantilever beam. The cantilever beam is a typical example to study bending problems~\cite{c18}. Standard analytical methods could quickly analyze the approximated structure, and we can derive the system parameters of dynamic model such as natural frequency. Damping ratio becomes the only remaining parameter which can be easily obtained by fitting step responses of the soft actuator.

The simplified structure of the soft actuator is shown in Fig.~\ref{fig: 2} (a). Before we analyze the approximated beam structure, we assume that the soft pneumatic actuator exhibits linear deformations. Usually, the soft materials show a nonlinear stress-strain relationship. However, if their deformations are not large, they still show a linear relationship. The stress-strain curve of soft silicone rubbers used in this research, Ecoflex Dragon Skin, can be approximated as a linear curve at strain around 100\%~\cite{c21}.

We model the soft actuator (simplified beam) as the second-order system to describe the dynamics of bending angle. The validity of the second-order system is analyzed in Sec.~\ref{robustness}. The standard form of the second-order model of the system is expressed as 
\begin{align}
\ddot{\theta}~+~\frac{C}{M}\dot{\theta}~+~\frac{K}{M}{\theta} = \frac{F}{M}
\label{eqn: 1}
\end{align}
\begin{figure}[http]
    \centering
    \vspace{0.1in}
    \includegraphics[width=200pt]{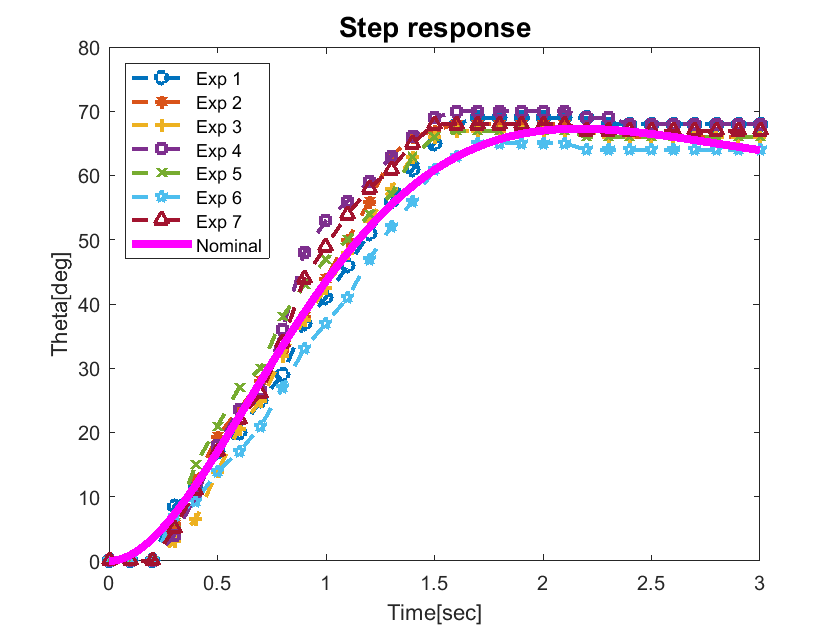}
    \caption{Step responses of the soft pneumatic actuator and the damping ratio and its perturbation term are obtained by fitting the nominal and deviated responses.}
    \vspace{-0.15in}
    \label{fig: 3}
\end{figure}
where $M$ is the mass of the soft actuator, $C$ is the damping constant of the system, $K$ is the spring constant of the system, and $F$ is the force acted at the beam generated by the input pressure. The $\frac{C}{M}$ and $\frac{K}{M}$ can also be expressed as $2{\zeta}{\omega_n}$ and ${\omega_n}^2$ respectively
\begin{align}
\ddot{\theta}~+~2{\zeta}{\omega_n}\dot{\theta}~+~{\omega_n}^2{\theta} = {F}/{M}
\label{eqn: 2}
\end{align}

The analysis can be observed in Fig.~\ref{fig: 2} (b). When a force is generated by input pressure, the beam structure deflects and has a bending angle $\theta$. The static equilibrium bending angle can be described as~\cite{c22}
\begin{align}
{\theta} = \frac{FL^2}{2EI}
\label{eqn: 3}
\end{align}
where $\theta$ is the bending angle of the approximated beam, $F$ is the force acting at the end generated by input pressure, $L$ is the length of the actuator, $E$ is Young's modulus, and $I$ is the moment inertia of the approximated beam. We manipulate the~(\ref{eqn: 3}) and obtain
\begin{align}
\frac{F}{\theta} = \frac{2EI}{L^2} = {K}
\label{eqn: 4}
\end{align}
where $K$ is the equivalent spring constant of the approximated beam structure under bending force. Hence, the spring constant of~(\ref{eqn: 1}) is obtained. Temporarily, the damping term is ignored. Then, the~(\ref{eqn: 1}) can be written as 
\begin{align}
\ddot{\theta}~+~\frac{2EI}{ML^2}{\theta} = \frac{F}{M}
\label{eqn: 5}
\end{align}

\noindent The $\frac{2EI}{ML^2}$ is the square of the natural frequency as the Eq. (\ref{eqn: 2}), and the natural frequency is
\begin{align}
\omega_n = \sqrt{\frac{2EI}{ML^2}}
\label{eqn: 6}
\end{align}

Next, the damping ratio is estimated by fitting the step response of the soft pneumatic actuator with a step input pressure (0.06 $MPa$) as in Fig.~\ref{fig: 3}. Due to the nonlinearity and unpredictability of the soft materials, the damping ratio is not a constant, but it comes with a perturbation term. The damping ratio of our soft actuators and its perturbation is around ${\zeta}+{\Delta\zeta}$ = $0.6\pm0.1$. Note that, the perturbed parameter introduces uncertainty to the model. However, the stability and robustness of the modeling approach are maintained, as discussed in Section~\ref{robustness}.

\begin{figure}[http]
    \centering
    \vspace{0.1in}
    \includegraphics[width=0.9\columnwidth]{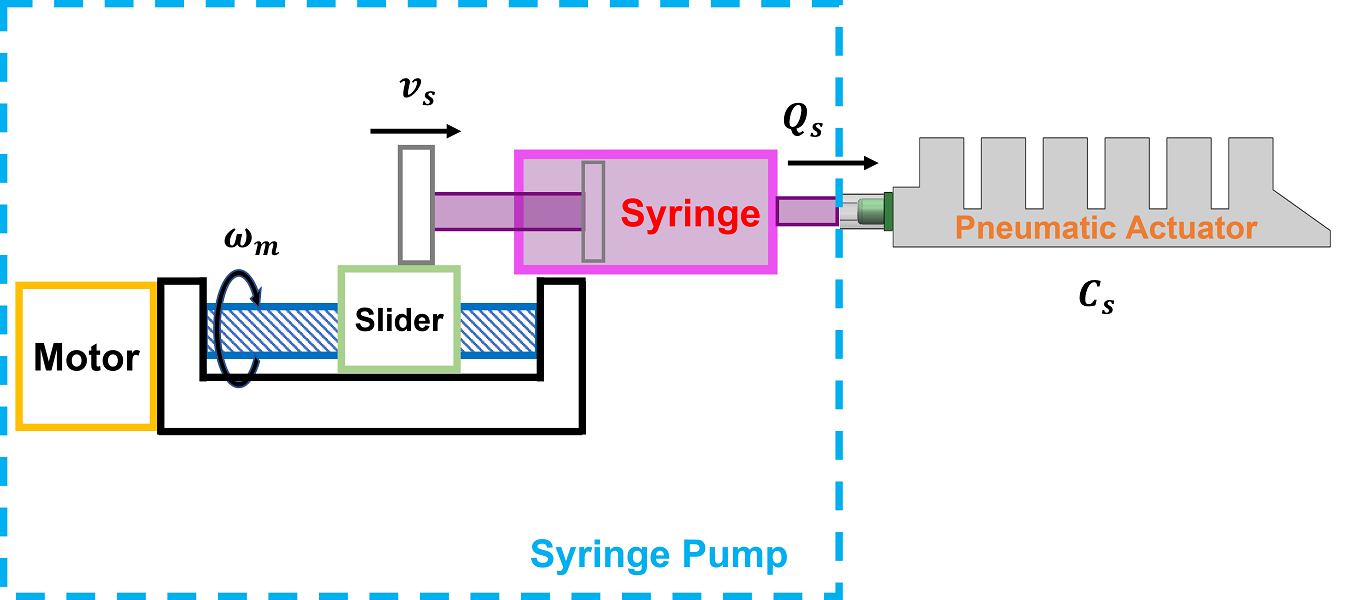}
    \caption{The schematic of the syringe pump which includes a linear actuator and a commercial syringe.}
    \vspace{-0.15in}

    \label{fig: 4}
\end{figure}

\subsection{Design and Modeling of the Syringe Pump}
\label{model}

Recent works usually utilize the syringe pump~\cite{syringe,s1} or air pumps with a pressure regulator or solenoid valve to generate air pressure for the pneumatic actuators and soft robots~\cite{c7, c27, c28}. The air pumps are easily accessible but they are low in resolution and usually need extra bulky reservoirs. Instead of using an air pump, we build a syringe pump to provide differential air pressure change for the pneumatic actuator. The pump does not need an air regulator. Instead, it simplifies the pressure control to motor control and can provide more precise air pressure. Moreover, the pressure dynamic model of the syringe pump is proposed as follows 

The schematic of the pressure control platform is displayed in Fig.~\ref{fig: 4}. A syringe could store air like the tank of the air pump. The slider of the linear actuator is connected to the syringe to push/pull and control the air pressure inside the soft actuator. The modeling of the pneumatic control platform is introduced in~\cite{c35}. The process begins with the relationship between motor speed and slider speed
\begin{align}
    {v_s} = \frac{l}{2\pi} {\omega_m}
\label{eqn: 7}
\end{align}
where $v_s$ is the speed of the slider in the linear actuator, $l$ is the lead of the screw in the linear actuator, and $\omega_m$ is the speed of the motor inside the linear actuator. The air output flow rate of the syringe is described as
\begin{align}
{Q_s} = {A_s} {v_s} = \frac{lA_s}{2\pi} {\omega_m}
\label{eqn: 8}
\end{align}
where ${Q_s}$ is the air output flow rate of the syringe, and $A_s$ is the inside cross-sectional area of the syringe. Lastly, the pressure changing rate inside the soft actuator is attained by dividing $Q_s$(Eq.~(\ref{eqn: 8})) by the capacity of soft actuator $C_s$
\begin{align}
\dot{P} = \frac{Q_s}{C_s} = \frac{l A_s}{2 \pi C_s} {\omega_m}
\label{eqn: 9}
\end{align}

\noindent Note that the capacity of the soft actuator will change as it is pressurized. However, we treat it as a constant, partly because the maximum expansion is still in the linear deformation range during operations, and partly because we would simplify the model and let the controller handle it. Thus, it is assumed as a constant in the modeling stage.

\subsection{Full System Model}

The transfer function of the system can be obtained by taking the Laplace transform of (\ref{eqn: 2}). Since we make a linear model assumption, $F=c\times P$ ($c$ is a constant) as in~\cite{c31}.

\begin{align}
{T_{SPA}} = \frac{cP/M}{s^2 + 2{\zeta}{\omega_n}s + {\omega_n}^2} 
\label{eqn: 10}
\end{align}
The transfer function of the pneumatic control system is obtained by taking the Laplace transform of (\ref{eqn: 9})

\begin{align}
{T_{PCS}} = \frac{lA_s{\omega_m}}{2\pi{C_s}}\frac{1}{s}
\label{eqn: 11}
\end{align}
We get the full system's transfer function by replacing $P$ in (\ref{eqn: 10}) by (\ref{eqn: 11}). 

\begin{align}
{T_{SYS}} = \frac{lA_s{\omega_m}c/2\pi{C_s}M}{s^3 + 2{\zeta}{\omega_n}s^2 + {\omega_n}^2s} 
\label{eqn: 12}
\end{align}
The full system is a third-order system.

\subsection{Uncertain Model Evaluation}
\label{robustness}
Mathematical models aim to represent reality by mapping inputs to responses. Yet, the inherent differences between mathematical representations and physical systems mean no model can be entirely accurate. This necessitates a level of trust among those employing these models. The challenge is crafting a model that is simple enough for practical use but detailed enough to be a reliable representation of reality.

While physical principles allow us to create analytical representations of certain systems, refining these representations with experimental data is essential for enhancing their accuracy and relevance. There are certain system dynamics that are either too complex to model with traditional physics or entirely elusive to such modeling efforts. This complexity underscores that relying on a singular nominal model may fall short of capturing the intricacies of physical systems. Therefore, by increasing the number of repeatable experiments and collected data, we can better approximate the true system behavior \cite{bayrak2022new}. By establishing a broader set of nominal dynamic models (or all possible plant family notions) based on this data, we increase our chances of representing real-world conditions more effectively.

\begin{figure}
\centering
\vspace{0.1in}
\includegraphics[width=0.8\columnwidth]{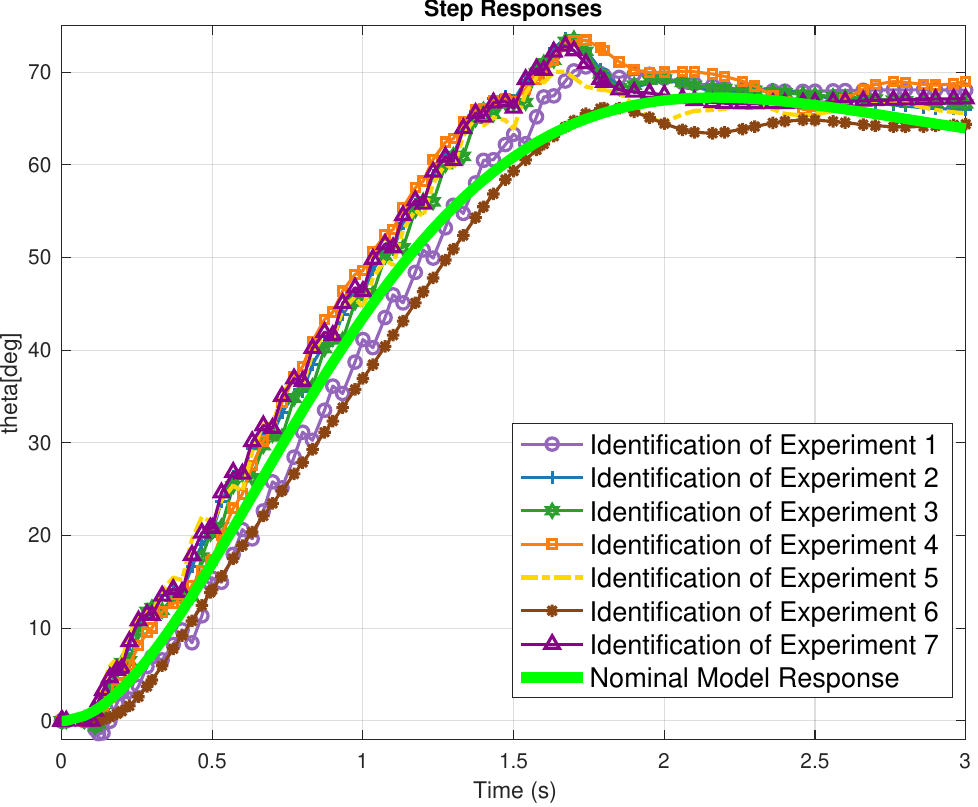}
\caption{Step responses of the fitted models in Table I.}
\vspace{-0.1in}

\label{f:bk1}
\end{figure}

\begin{figure}
\centering
\vspace{0.1in}
\includegraphics[width=0.8\columnwidth]{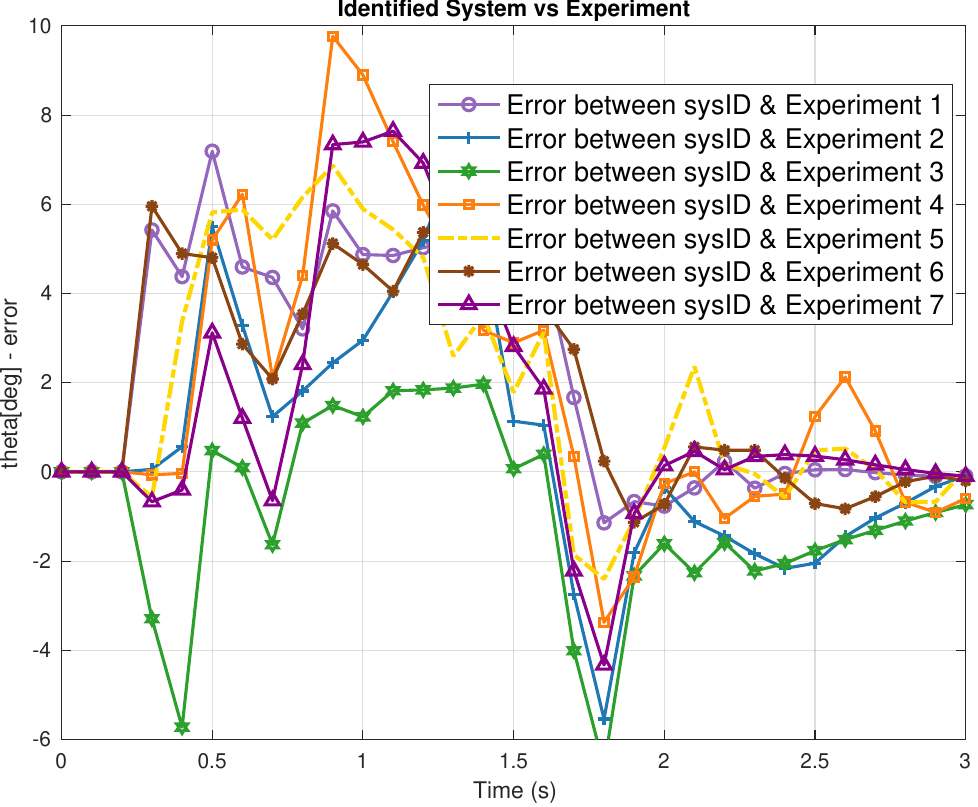}
\caption{The errors between step responses of the fitted models in Table I and the experimental step responses in Fig. 3.}
\vspace{-0.1in}

\label{f:bk2}
\end{figure}

To substantiate our approach, we executed seven distinct experiments, depicted in Fig. \ref{fig: 3}. The results show that, under almost identical conditions, each open-loop experiment yielded different outputs from the same setup. This indicates potential gaps in our model parameters or that system dynamics varying based on unidentified factors. To establish a comprehensive "all possible plant family", we initiate a series of system identification processes for each experiment. During this identification, we'll apply the N4SID technique, correlating identical inputs to their varied outputs as Fig.~\ref{f:bk1}. The efficacy of each system identification is detailed in Table I, while the discrepancies between each experimental result and its identified model are illustrated in Fig. \ref{f:bk2}.

\begin{table}[ht]
\caption{Fit to estimation data for each experiment in Fig. 3.}
\label{tab:my_label}
\centering
\begin{tabular}{|c|c|c|}
\hline
\textbf{Experiment} & \textbf{State-Space Order} & \textbf{Fit to Estimation Data (\%)} \\
\hline
P1 & 5 & 97.07 \\
\hline
P2 & 5 & 94.63 \\
\hline
P3 & 4 & 96.05 \\
\hline
P4 & 6 & 96.88 \\
\hline
P5 & 6 & 96.44 \\
\hline
P6 & 2 & 96.01 \\
\hline
P7 & 4 & 93.61 \\
\hline
\end{tabular}
\end{table}

The concept of the "all possible plant family" defined by $\hat{T}$ is expressed as:
\begin{equation} \label{eq:3}
\hat{T} \in \left\{ T(1+\Delta W_T) \quad | \quad \forall \ \|\Delta\|_\infty \leq 1 \right\}
\end{equation}
where
$W_T$ is a robustness weight function and $\Delta$ is any stable unstructured uncertainty function. A generic way to describe the robustness weight function $W_T$ is given in \cite{kurkccu2018disturbance} as
\begin{align}\label{WT_procedure}
\left| \frac{M_{ik}e^{j\phi_{ik}}}{M_{i}e^{j\phi_{i}}}-1\right| \leq \left| W_T(j\omega_i)\right|, i=1,\ldots,m;k=1,\ldots,n
\end{align}
Here, the magnitude and phase are evaluated across a set of frequencies, represented as $\omega_i$ (spanning from $i=1,\ldots,m$), and the experiment is reiterated $n=7$ times. ($M_{ik},\phi_{ik}$) delineate the magnitude-phase pair measurements corresponding to frequency $\omega_i$ and the experiment iteration $k$. ($M_{i},\phi_{i}$) denotes the magnitude-phase pairs for the nominal plant $T$.  The findings derived from adopting this methodology are given in Fig. \ref{f:bk1} with a harmony of Fig. \ref{fig: 3}. In our evaluation process for the choice of the nominal model, we find evidence, as illustrated in Fig. \ref{f:bk1}, that the analytical system model derived from \eqref{eqn: 10} aligns with the $\hat{T}$ family. This alignment not only reinforces the rationality behind our model selection but also underscores its utility. Adopting this nominal choice streamlines subsequent phases of design and analysis, allowing for a cohesive and consistent approach

\begin{figure}
\centering
\vspace{0.1in}
\includegraphics[width=0.8\columnwidth]{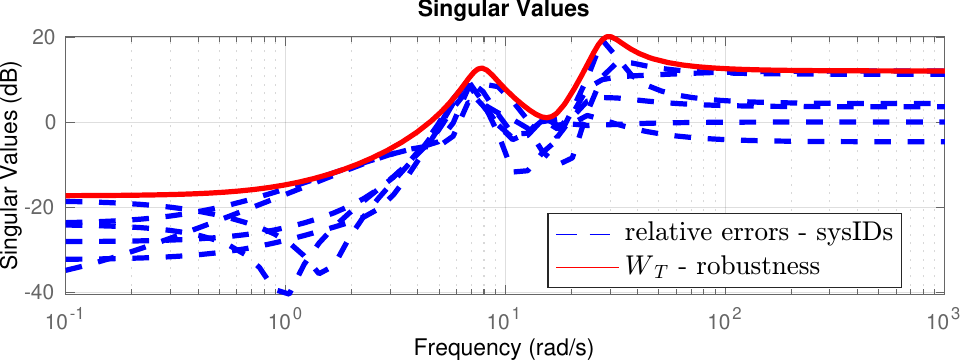}
\caption{Robustness weight selection based on the relative modeling errors.}
\vspace{-0.1in}

\label{f:bk3}
\end{figure}

Finally, if the designed controller satisfies $\Vert W_TTK(1+TK)^{-1} \Vert_\infty<1$, where $K$ stands for the controller in a compact form, the closed-loop system remains stable under all possible perturbations as in Fig. \ref{f:bk3}.

\section{Controller Design}
\label{controller}
This section aims to design a suitable controller for the soft pneumatic actuators. We begin with the open-loop analysis and then design the LQR for the soft actuators. 

\subsection{Open Loop Analysis}

With the system model (\ref{eqn: 12}), we can perform open-loop analysis by examining the open-loop poles which are ${0}$, ${-{\zeta}{\omega_n}\pm j{\omega_n}\sqrt{1 - {\zeta}^2}}$. There are one pole on the imaginary axis and two poles on the left half-plane if $-{\zeta}{\omega_n} < 0$. Thus, the system is marginally stable. 

The system does not have the resonance at the corner frequency and it begins to show phase lag in low frequency due to the soft materials.

\subsection{LQR Controller Design}
\label{LQR}
A PID controller was designed to regulate the soft actuator in our previous work~\cite{c35} with an uncompensated steady-state error. Thus, to address this steady-state error, meet the robustness condition, and enhance the performance, we have opted for an LQR controller to enhance performance. 
The LQR controller design seeks to minimize a cost function, \( J \), representing a balance between state performance and control effort: $J = \int_{0}^{\infty} ({\bf x}^T(t) Q {\bf x}(t) + {\bf u}^T(t) R {\bf u}(t)) \, dt 
$
Here, \( {\bf x}^T(t) Q {\bf x}(t) \) weighs the state deviations, while \( {\bf u}^T(t) R {\bf u}(t) \) assesses the control effort. By selecting appropriate \( Q \) and \( R \) matrices, the LQR fine-tunes this trade-off to achieve desired system performance with optimal control effort.

At first, the state vector is defined as $x=[\theta~\dot{\theta}~ \ddot{\theta}]^T$, and the system model (\ref{eqn: 12}) is formulated as the controllable canonical form as

\begin{align}
{\bf \dot x} =  A {\bf x} + B {\bf u}, \ {\bf y} = C {\bf x}
\label{eqn: 13}
\end{align}

where
\begin{align}
A = \begin{bmatrix} 0 & 1 & 0\\0 & 0 & 1\\ 0 & -{\omega_n}^2 & -2{\zeta}{\omega_n} \end{bmatrix},
B = \begin{bmatrix} 0 \\0 \\ 1\end{bmatrix},
C =\begin{bmatrix} \frac{lA_s{\omega_m}c}{2\pi{C_s}M} \\ 0 \\ 0\end{bmatrix}^T
\label{eqn: 15}
\end{align}

The LQR algorithm is an automated method of finding a suitable state-feedback controller $(u)$, which is attained by solving the continuous-time algebraic Riccati equation~\cite{c30}

\begin{align}
{A^T}Y + Y{A} + YB{R^{-1}}{B^T}Y + Q = 0
\label{eqn: 16}
\end{align}
where $R=1$ and 

\begin{align}
Q = p\begin{bmatrix} 1 & 0 & 0\\0 & 0.1 & 0\\ 0 & 0 & 0 \end{bmatrix}
\label{eqn: 17}
\end{align}
where we penalize the bending angle $\theta$ and its velocity $\dot \theta$, and $p\in R$. Once we get the $Y$ by solving (\ref{eqn: 16}), we have the state-feedback controller
\begin{align}
{u} = -R^{-1}{B^T}Y{\bf x}
\label{eqn: 18}
\end{align}

\noindent The stability of the system Eq. (\ref{eqn: 13}) can be checked by defining a Lyapunov function
\begin{align}
{V} =  {\bf x}^T{Y}{\bf x}
\label{eqn: 19}
\end{align}
\begin{align}
\dot{V} =  \dot{\bf x}^T{Y}{\bf x} + {\bf x}^T{Y}\dot{\bf x} = {\bf x}^T({A^T{Y}+YA}){\bf x}
\label{eqn: 20}
\end{align}
\noindent Since the $Y > 0$, the Lyapunov function $V$ is positive definite and decrescent. The $-\dot{V}$ is positive definite. Hence, the system is uniformly globally stable based on the Lyapunov stability criterion~\cite{c33}. 

\subsection{Simulation Results}

Simulations are conducted in MATLAB/Simulink to evaluate the performance of the controllers designed in Sec.~\ref{LQR} and numerically validate the robustness criteria derived in Sec.~\ref{robustness}. The reference is the step function which is set as $\theta = \pi/2$. The simulations are conducted in two different soft actuators. In addition, we intend to design a controller that makes the soft actuators respond faster and reach states precisely, so the selected indices are the settling time and steady-state error. The smaller the settling time and steady-state error, the better the controller. 
\begin{figure}[http]
    \centering
    \includegraphics[width=200pt]{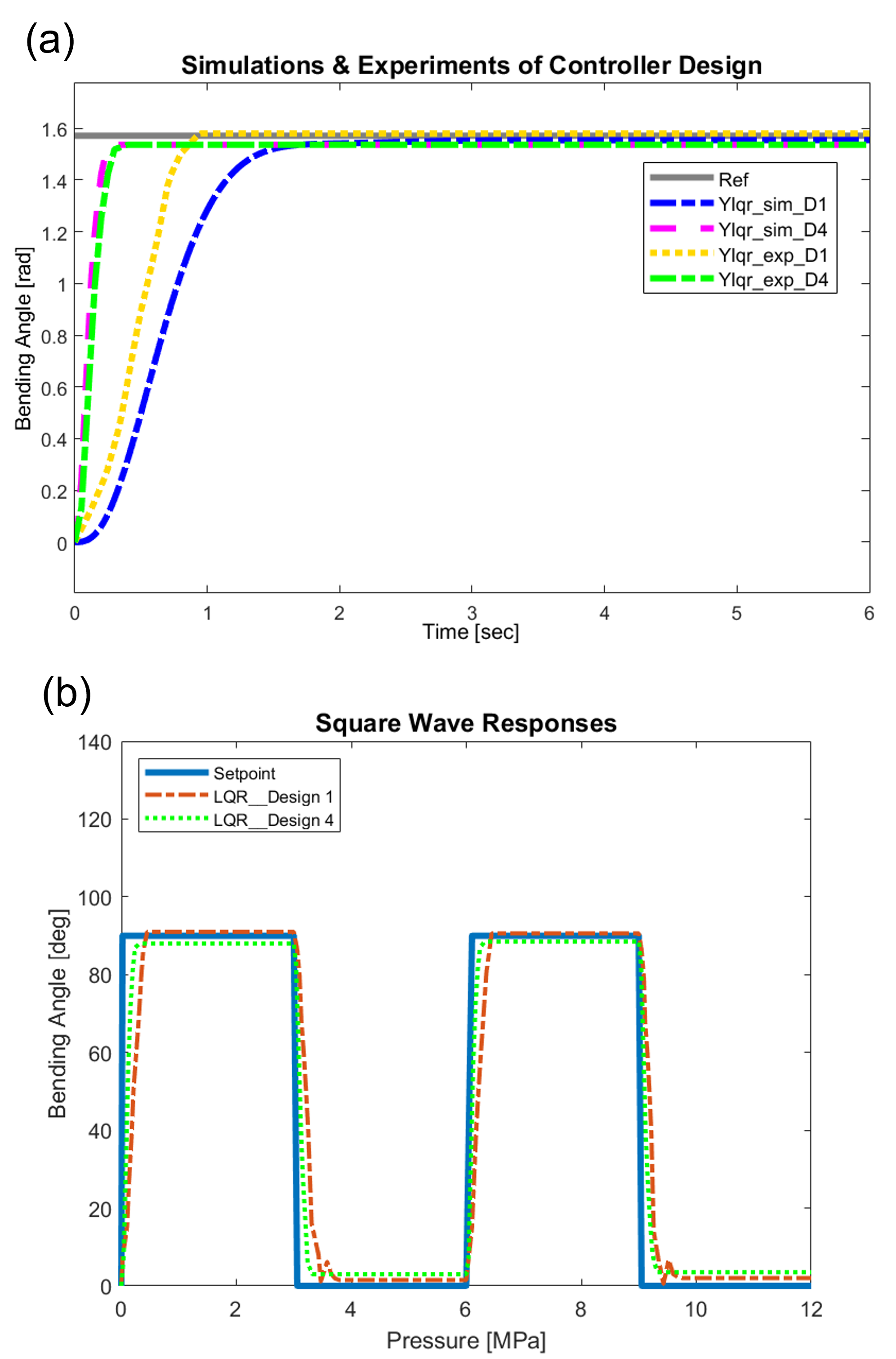}
    \caption{(a) The simulations and experimental validation of LQR controller designed for the soft actuators (Design 1 and Design 4). (b) The square wave function responses of Design 1 and Design 4 are demonstrated. }
    \vspace{-0.15in}
    \label{fig: 9}
\end{figure}

The soft actuators were optimal designs in our previous work~\cite{c3, c31}. The former work relied on the data-driven method to search for the optimal parameters, while the later one implemented the model-based optimization approach to explore the optimal parameters. The prototypes are used to verify the modeling and controller design method. The simulation results are demonstrated in Fig.~\ref{fig: 9} (a). Their dynamical models are obtained by getting $\zeta$ and $\omega_n$ discussed in Sec. II. In Fig.~\ref{fig: 9}(a), Design 1 is developed in~\cite{c31} which is made of a material whose Young's modulus is 0.34 MPa~\cite{c34}. 
The LQR controller reaches a settling time of around 0.8 seconds and nearly cancel the steady-state error by penalizing the state $\theta$ and $\dot \theta$. Design 4 is designed in~\cite{c3} which is made of stiffer material whose Young's modulus is around 10 MPa determined by tensile test. Its settling time is approximately 0.5 seconds as in Fig.~\ref{fig: 9}(a).

\section{Experimental Evaluation}
\subsection{Experimental Setup}

\begin{figure}[http]
    \centering
    \vspace{0.1in}
    \includegraphics[width=200pt]{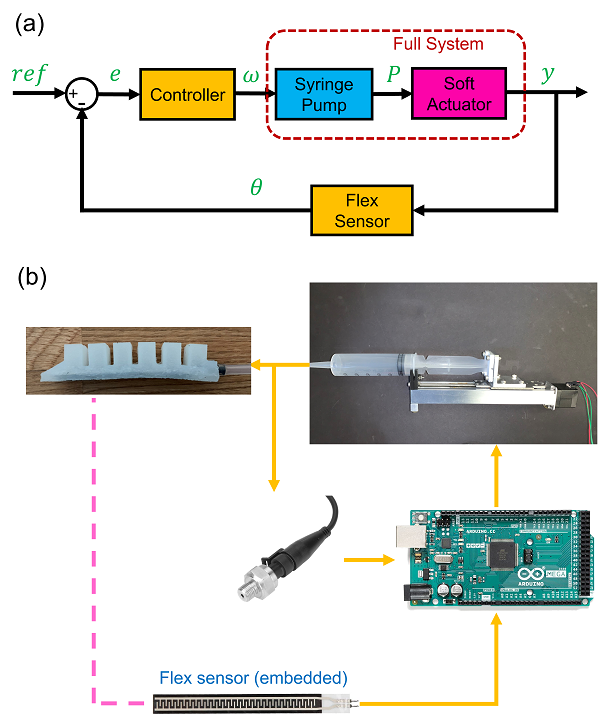}
    \caption{(a) The control block diagram in both MATLAB/Simulink simulations and experiments. (b) The schematic of the experimental setup and signal flows.}
    \vspace{-0.15in}
    \label{fig: 8}
\end{figure}

The control block diagram and the experimental setup are displayed in Fig.~\ref{fig: 8}. Soft actuators are driven by the self-designed syringe pump. An air pressure sensor (Walfront, Lewes, DE) with a sensing range of 0 to 80~psi is utilized to detect the air pressure for open-loop control. A flex sensor (Walfront, Lewes, DE) is embedded inside the actuator to observe the bending angle for feedback control. Both sensors are synchronized with Arduino MEGA 2560 (SparkFun Electronics, Niwot, CO). The microcontroller is based on the Microchip ATmega 2560. 
The microcontroller is also synchronized with a computer to log sensing data.

\begin{table}[http] 
\setlength{\tabcolsep}{1pt}
\centering
\caption{\label{tab:Table II}Comparisons of true natural frequencies and the estimations from the derived equation}
\begin{tabular}{|c c c c|}
\hline
Unit [rad/s] & True $\omega_n$ & Estimated $\omega_n$ & Error\\ [0.5ex]
\hline
\makecell{Design 1~\cite{c31} \\(E=0.34MPa, M=0.17N, L=0.94m)} & 1.900$\pm$0.035  & 1.812 & 4.86\%\\ [0.5ex] 
\hline
\makecell{Design 2~\cite{c31} \\(E=0.26MPa, M=0.24N, L=0.94m)} & 1.141$\pm$0.046 & 1.372 &  16.84\%\\ [0.5ex]
\hline
\makecell{Design 3~\cite{c31} \\(E=0.34MPa, M=0.20N, L=0.106m)} & 1.523$\pm$0.046 & 1.422 & 7.10\% \\ [0.5ex] 
\hline
\makecell{Design 4~\cite{c3} \\(E=10MPa, M=0.04N, L=0.060m)} & 10.420$\pm$0.094  & 8.709 & 19.64\%\\ [0.5ex] 
\hline
\end{tabular}
\vspace{-0.15in}

\end{table}

\subsection{Single Soft Actuator Test}
\subsubsection{Verification of Dynamical Model}
In Sec.~\ref{model}, we build the second-order dynamical model for a soft actuator. The damping ratio $\zeta$ is obtained by fitting the true responses of the soft actuator, and the natural frequency is determined by Eq.~(\ref{eqn: 6}). Thus, this sub-section aims to verify the accuracy of the natural frequency by comparing the true natural frequencies and estimations of the equation. Based on the Eq.~(\ref{eqn: 6}), the natural frequency is influenced by Young's modulus ($E$), moment of inertia ($I$), mass of actuator ($M$), and its length ($L$). Since the moment of inertia $I$ of the actuators are decided during the design stage in our previous works~\cite{c3, c31}, so they are fixed here. We use different materials ($E$) and different lengths ($L$), and both different materials and lengths will affect the mass ($M$). The comparison results are illustrated in Table II. 

The error of the Eq.~(\ref{eqn: 6}) ranges from approximately 4.86 \% to 19.64 \%, which is influenced by the $E$ and $L$. Design 1 to Design 3 have the same shape, height, and width but have different lengths or materials. Design 1 has the smallest error 4.86 \%. The longer the actuator and the smaller Young's modulus, the larger the errors of the Eq.~(\ref{eqn: 6}). For example, Design 2 has an error of 16.84 \% because it is made of softer material. Design 4 is made of harder material, has different dimensional parameters, and has a quite distinct geometric shape. The error is increased to about 19.64 \%. 

\subsubsection{Step responses}
Step response is the time-domain behavior of the output of a system with a step-input command. Step response test helps gauge the performance of the controller such as the steady-state error and the settling time. The step responses of LQR controllers of Design 1 are displayed in Fig.~\ref{fig: 9} (a). The experimental results of the LQR controller respond faster than the simulation result. The step response has nearly zero steady-state error and the settling time is around 0.8 seconds. The result of Design 4 is shown in Fig.~\ref{fig: 9} (a). The performance is better due to the property of stiffer material. The settling time is around 0.5 seconds. 

In addition, we also test the step responses of the actuators with reference 90 degrees (like Fig.~\ref{fig: 9}(a)) in Table II 10 times to get the averages and standard deviations as displayed in Table III. All actuators are controlled by the LQR controller. Again, Design 2 (softer material) and Design 3 (longer structure) have relatively larger errors and standard deviations. Design 4 has a relatively smaller error and standard deviation compared to Design 1.

\begin{table}[http] 
\centering
\caption{\label{tab:Table III}The average of step responses and standard deviation of the soft actuators in Table II}
\begin{tabular}{|c c c|}
\hline
& Average [deg] & STD [deg]\\ [0.5ex]
\hline
Design 1~\cite{c31} & 91.1  & 1.13\\ [0.5ex] 
\hline
Design 2~\cite{c31} & 92.1 & 4.09 \\ [0.5ex]
\hline
Design 3~\cite{c31} & 91.9 & 2.23 \\ [0.5ex] 
\hline
Design 4~\cite{c3} & 90.5  & 0.76 \\ [0.5ex] 
\hline
\end{tabular}

\end{table}

\subsubsection{Square Wave Responses}
The square wave response is to know the performance of the soft actuator with continuously changing input command. We can observe the soft actuator's errors and delay of responses. The square wave responses are demonstrated in Fig.~\ref{fig: 9} (b). The Red dashed line represents the result of Design 1 while the green dashed line describes the response of Design 4. Again, Design 1 shows a slightly delay response compared to Design 4 due to the material property. Also, the true response has slight vibrations when the soft actuator is returning to the 0 degrees because of the asymmetric structure of the soft actuator.

\subsection{Two-fingered Gripper Test}
\begin{figure}[http]
    \centering
    \vspace{0.1in}
    \includegraphics[width=210pt]{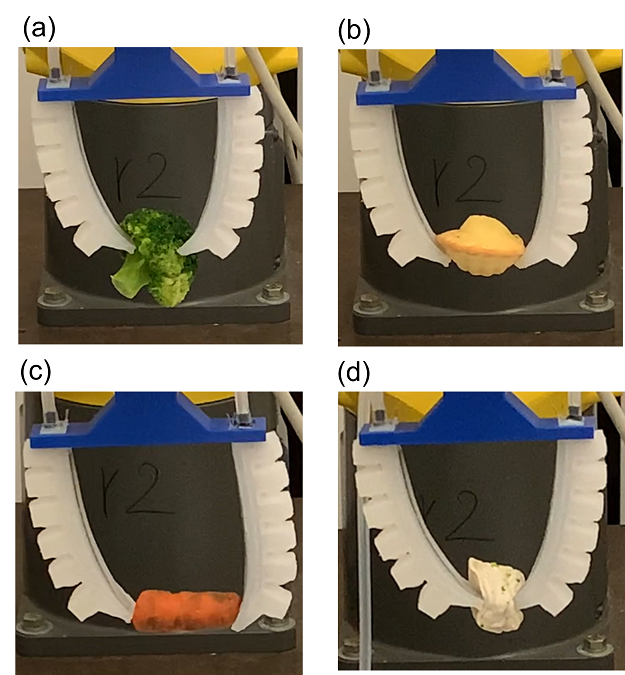}
    \caption{The soft gripper with feedback controller is deployed on an industrial robot arm to grasp various food such as cauliflower, bread, carrot, and dumpling.}
    \vspace{-0.2in}
    \label{fig: 11}
\end{figure}

Two soft actuators of Design 1s~\cite{c31} are used to make a two-fingered soft gripper. Usually, the open-loop controller cannot synchronize the multiple fingers(different response times) to hold an object stably stage, so we implement the full system model and the LQR controller designed in this research to address this issue. The experimental results are demonstrated in Fig.~\ref{fig: 11}. The gripper is used to grasp food and vegetables. The asynchronized response times of different fingers caused by the open-loop control would weaken the manipulability of the soft gripper, and the objects tend to drop. By contrast, the synchronized motions of two fingers regulated by the feedback controller are able to grasp the objects successfully. Thus, we control the robot to grasp four different objects for 10 times. The total success rate of the gripper with open-loop control is 67.5\%, while the success rate of the gripper with closed-loop control is increased to 92.5\%. 

\section{Conclusion and Future Works}

This work presents a full system modeling approach and control strategy for a soft pneumatic actuator driven by the syringe pump. The soft actuator is modeled based on Euler’s bending theory. The analytical model of natural frequency is proposed. The damping ratio is estimated by curve fitting step responses of actuators. This modeling approach is validated on four soft actuators designed by both model-based and data-driven methods. The modelling error can be as low as 4.86 \%. Moreover, the modeling method is proved to be stable and robust under model uncertainty evaluation. Furthermore, the pressure dynamic model of the syringe pump is derived to build the full system model(models of syringe pump and soft actuator). The optimal controller(LQR) is implemented to control the full system. The LQR controller enables the system to achieve fast responses and higher accuracy. Given the desired reference(90 degrees), the settling time is reduced to 0.8 seconds from about 2.2 seconds and the standard deviation of steady-state error is approximately 1 degree. The full system with LQR controller also performs well in square wave function test, and the experimental results are as good as the model simulations in MATLAB environment. Finally, the coordinated control of the gripper's fingers imparts stability and precision, enabling the successful and secure grasping of diverse delicate objects, augmenting the success rate of gripping operations.

Our future works include nonlinear parameter-varying modeling and underactuated control for soft pneumatic actuators, using the soft pneumatic actuators for grasping~\cite{zhu2021grasp} tasks, and developing algorithms to apply the hardware to the multiple tasks in real world using multi-task RL~\cite{sun2022mtrl}. 

\section*{Acknowledgement}
The authors would like to thank NSK, Ltd. for arranging the linear actuators used in the experiments.

\bibliographystyle{ieeetr}
\bibliography{IEEEabrv}

\begin{thebibliography}{10}

\bibitem{c1}
F.~Iida and C.~Laschi, ``Soft robotics: Challenges and perspectives,'' {\em Procedia Computer Science}, vol.~7, no.~1, pp.~99--102, 2011.

\bibitem{c2}
W.-T. Yang and M.~Tomizuka, ``Design a multifunctional soft tactile sensor enhanced by machine learning approaches,'' {\em ASME Journal of Dynamic Systems, Measurement, and Control}, vol.~144, no.~8, p.~081006, 2022.

\bibitem{c3}
K.~G. Demir, Z.~Zhang, J.~Yang, and G.~X. Gu, ``Computational and experimental design exploration of 3d‐printed soft pneumatic actuators,'' {\em Advanced Intelligent Systems}, vol.~2, p.~7, 2020.

\bibitem{c4}
E.~Navas, R.~Fernández, D.~Sepúlveda, M.~Armada, and P.~Gonzalez-de Santos, ``Soft grippers for automatic crop harvesting: A review,'' {\em Sensors}, vol.~21, no.~8, p.~2689, 2021.

\bibitem{c25}
T.~George~Thuruthel, Y.~Ansari, E.~Falotico, and C.~Laschi, ``Wang, zhongkui and or, keung and hirai, shinichi,'' {\em Robotics and Autonomous Systems}, vol.~125, p.~103427, 2020.

\bibitem{c5}
S.~Zaidi, M.~Maselli, C.~Laschi, and M.~Cianchetti, ``Actuation technologies for soft robot grippers and manipulators: A review,'' {\em Current Robotics Reports}, pp.~1--15, 2021.

\bibitem{zhang2022efficient}
Z.~Zhang, Z.~Jin, and G.~X. Gu, ``Efficient pneumatic actuation modeling using hybrid physics-based and data-driven framework,'' {\em Cell Reports Physical Science}, vol.~3, no.~4, p.~100842, 2022.

\bibitem{c6}
J.~Hughes, U.~Culha, F.~Giardina, F.~Guenther, A.~Rosendo, and F.~Iida, ``Soft manipulators and grippers: a review,'' {\em Frontiers in Robotics and AI}, vol.~3, p.~69, 2016.

\bibitem{c7}
G.~Cao, B.~Huo, L.~Yang, F.~Zhang, Y.~Liu, and G.~Bian, ``Model-based robust tracking control without observers for soft bending actuators,'' {\em IEEE Robotics and Automation Letters}, vol.~6, no.~3, pp.~5175--5182, 2021.

\bibitem{c8}
M.~S. Xavier, A.~J. Fleming, and Y.~K. Yong, ``Nonlinear estimation and control of bending soft pneumatic actuators using feedback linearization and ukf,'' {\em IEEE/ASME Transactions on Mechatronics}, 2022.

\bibitem{c26}
T.~George~Thuruthel, Y.~Ansari, E.~Falotico, and C.~Laschi, ``Control strategies for soft robotic manipulators: A survey,'' {\em Soft robotics}, vol.~5, no.~2, pp.~149--163, 2018.

\bibitem{c14}
E.~H. Skorina, M.~Luo, W.~Tao, F.~Chen, J.~Fu, and C.~D. Onal, ``Adapting to flexibility: model reference adaptive control of soft bending actuators,'' {\em IEEE Robotics and Automation Letters}, vol.~2, no.~2, pp.~964--970, 2017.

\bibitem{c15}
T.~Wang, Y.~Zhang, Z.~Chen, and S.~Zhu, ``Parameter identification and model-based nonlinear robust control of fluidic soft bending actuators,'' {\em IEEE/ASME transactions on mechatronics}, vol.~24, no.~3, pp.~1346--1355, 2019.

\bibitem{c18}
E.~H. Skorina, M.~Luo, S.~Ozel, F.~Chen, W.~Tao, and C.~D. Onal, ``Feedforward augmented sliding mode motion control of antagonistic soft pneumatic actuators,'' in {\em 2015 IEEE International Conference on Robotics and Automation (ICRA)}, pp.~2544--2549, IEEE, 2015.

\bibitem{c19}
E.~H. Skorina, W.~Tao, F.~Chen, M.~Luo, and C.~D. Onal, ``Motion control of a soft-actuated modular manipulator,'' in {\em 2016 IEEE International Conference on Robotics and Automation (ICRA)}, pp.~4997--5002, IEEE, 2016.

\bibitem{c20}
M.~d.~S. Xavier, A.~J. Fleming, and Y.~K.~K. Yong, ``Model-based nonlinear feedback controllers for pressure control of soft pneumatic actuators using on/off valves,'' {\em Frontiers in Robotics and AI}, p.~33, 2022.

\bibitem{c32}
Z.~Q. Tang, H.~L. Heung, K.~Y. Tong, and Z.~Li, ``Model-based online learning and adaptive control for a “human-wearable soft robot” integrated system,'' {\em The International Journal of Robotics Research}, vol.~40, no.~1, pp.~256--276, 2021.

\bibitem{c22}
R.~C. Hibbeler, {\em Mechanics of materials 8th}.
\newblock Pearson, New York, 2017.

\bibitem{c9}
C.~Della~Santina, R.~K. Katzschmann, A.~Biechi, and D.~Rus, ``Dynamic control of soft robots interacting with the environment,'' in {\em IEEE International Conference on Soft Robotics (RoboSoft)}, pp.~46--53, IEEE, 2018.

\bibitem{c10}
K.~M. de~Payrebrune and O.~M. O’Reilly, ``On constitutive relations for a rod-based model of a pneu-net bending actuator,'' {\em Extreme Mechanics Letters}, vol.~9, pp.~38--46, 2016.

\bibitem{c11}
A.~Doroudchi and S.~Berman, ``Configuration tracking for soft continuum robotic arms using inverse dynamic control of a cosserat rod model,'' in {\em IEEE International Conference on Soft Robotics (RoboSoft)}, pp.~207--214, IEEE, 2021.

\bibitem{c24}
A.~S. Lafmejani, H.~Farivarnejad, A.~Doroudchi, and S.~Berman, ``A consensus strategy for decentralized kinematic control of multi-segment soft continuum robots,'' in {\em 2020 American Control Conference (ACC)}, pp.~909--916, IEEE, 2020.

\bibitem{c12}
C.~M. Best, M.~T. Gillespie, P.~Hyatt, L.~Rupert, V.~Sherrod, and M.~D. Killpack, ``A new soft robot control method: Using model predictive control for a pneumatically actuated humanoid,'' {\em IEEE Robotics and Automation Magazine}, vol.~23, no.~3, pp.~75--84, 2016.

\bibitem{c16}
T.~George~Thuruthel, F.~Renda, and F.~Iida, ``First-order dynamic modeling and control of soft robots,'' {\em Frontiers in Robotics and AI}, vol.~7, p.~95, 2019.

\bibitem{syringe}
T.~Kalisky, Y.~Wang, B.~Shih, D.~Drotman, S.~Jadhav, E.~Aronoff-Spencer, and M.~T. Tolley, ``Differential pressure control of 3d printed soft fluidic actuators,'' in {\em IEEE/RSJ International Conference on Intelligent Robots and Systems (IROS)}, pp.~6207--6213, IEEE, 2017.

\bibitem{s1}
M.~S. Xavier, A.~J. Fleming, and Y.~K. Yong, ``Image-guided locomotion of a pneumatic-driven peristaltic soft robot,'' in {\em 2019 IEEE International Conference on Robotics and Biomimetics (ROBIO)}, pp.~2269--2274, IEEE, 2019.

\bibitem{c13}
Z.~Wang and S.~Hirai, ``Soft gripper dynamics using a line-segment model with an optimization-based parameter identification method,'' {\em IEEE Robotics and Automation Letters}, vol.~2, no.~2, pp.~624--631, 2017.

\bibitem{c17}
A.~D. Marchese, K.~Komorowski, C.~D. Onal, and D.~Rus, ``Design and control of a soft and continuously deformable 2d robotic manipulation system,'' in {\em 2014 IEEE international conference on robotics and automation (ICRA)}, pp.~2189--2196, IEEE, 2014.

\bibitem{c27}
W.~Shi and M.~B. Wijesundara, ``Angular velocity control of pneumatic soft robotic digits,'' {\em ASME Letters in Dynamic Systems and Control}, vol.~1, no.~3, 2021.

\bibitem{c21}
G.~Belforte, F.~Dabbene, and P.~Gay, ``Sensor design, identification and control of a deformable pneumatic actuator,'' {\em International Journal of Mechanics and Control}, vol.~4, no.~1, pp.~3--13, 2003.

\bibitem{c23}
M.~Thieffry, A.~Kruszewski, C.~Duriez, and T.-M. Guerra, ``Control design for soft robots based on reduced-order model,'' {\em IEEE Robotics and Automation Letters}, vol.~4, no.~1, pp.~25--32, 2018.

\bibitem{c28}
M.~S. Xavier, A.~J. Fleming, and Y.~K. Yong, ``Design and control of pneumatic systems for soft robotics: A simulation approach,'' {\em IEEE Robotics and Automation Letters}, vol.~6, no.~3, pp.~5800--5807, 2021.

\bibitem{c35}
W.-T. Yang, M.~Hirao, and M.~Tomizuka, ``Design, modeling, and parametric analysis of a syringe pump for soft pneumatic actuators,'' in {\em 2023 IEEE/ASME International Conference on Advanced Intelligent Mechatronics (AIM)}, pp.~317--322, IEEE, 2023.

\bibitem{c31}
W.-T. Yang, H.~S. Stuart, and M.~Tomizuka, ``Mechanical modeling and optimal model-based design of a soft pneumatic actuator,'' in {\em 2023 IEEE International Conference on Soft Robotics (RoboSoft)}, IEEE, 2023 (accepted).

\bibitem{bayrak2022new}
A.~Bayrak, B.~K{\"u}rk{\c{c}}{\"u}, and M.~{\"O}. Efe, ``A new adaptive disturbance/uncertainty estimator based control scheme for lti systems,'' {\em IEEE Access}, vol.~10, pp.~106849--106858, 2022.

\bibitem{kurkccu2018disturbance}
B.~K{\"u}rk{\c{c}}{\"u}, C.~Kasnako{\u{g}}lu, and M.~{\"O}. Efe, ``Disturbance/uncertainty estimator based robust control of nonminimum phase systems,'' {\em IEEE/ASME Transactions on Mechatronics}, vol.~23, no.~4, pp.~1941--1951, 2018.

\bibitem{c30}
B.~D. Anderson and J.~B. Moore, {\em Optimal control: linear quadratic methods}.
\newblock Courier Corporation, 2007.

\bibitem{c33}
S.~Sastry, {\em Nonlinear systems: analysis, stability, and control}.
\newblock Springer Science and Business Media, 2013.

\bibitem{c34}
``{Ecoflex Dragon Skin 20}.'' \url{https://www.smooth-on.com/products/dragon-skin-20/}.
\newblock Accessed: 2022-06-07.

\bibitem{zhu2021grasp}
X.~Zhu, L.~Sun, Y.~Fan, and M.~Tomizuka, ``6-dof contrastive grasp proposal network,'' in {\em 2021 IEEE International Conference on Robotics and Automation (ICRA)}, pp.~6371--6377, 2021.

\bibitem{sun2022mtrl}
L.~Sun, H.~Zhang, W.~Xu, and M.~Tomizuka, ``Paco: Parameter-compositional multi-task reinforcement learning,'' in {\em NeurIPS}, 2022.

\end{thebibliography}

\end{document}